\RequirePackage{fix-cm}
\documentclass[smallextended]{svjour3}       
\smartqed  
\usepackage{graphicx}
\usepackage{bm}
\usepackage[utf8]{inputenc} 
\usepackage{url}            
\usepackage{booktabs}       
\usepackage{nicefrac}       
\usepackage{microtype}      
\usepackage{amsmath,amssymb,amsfonts}
\usepackage{multirow}
\usepackage{times}
\usepackage{latexsym}
\usepackage{array}
\usepackage{bm}
\usepackage{ulem}
\usepackage[ruled,vlined]{algorithm2e}
\usepackage{balance}
\usepackage{subfigure}
\usepackage{diagbox} 
\usepackage{float}
\usepackage{graphicx}
\usepackage{textcomp}
\usepackage{xcolor}
\usepackage{mathtools}

\newtheorem{prop}[theorem]{Proposition}    
\begin{document}

\title{Stein Variational Gradient Descent with Multiple Kernel
}


\author{Qingzhong Ai    \and
        Shiyu Liu   \and
        Lirong He   \and
        Zenglin Xu 
}

\institute{Qingzhong Ai, Shiyu Liu, Lirong He \at
              SMILE Lab, School of Computer Science and Engineering, University of Electronic Science and Technology of China, Chengdu, Sichuan, China
           \and
          Zenglin Xu (Corresponding author)\at
           School of Computer Science and Technology, Harbin Institute of Technology, Shenzhen, China. \\
           Center of Artificial Intelligence, Peng Cheng Lab, Shenzhen, Guangdong, China. \\
            \email{xuzenglin@hit.edu.cn}
}

\date{Received: date / Accepted: date}

\maketitle

\begin{abstract}
\textbf{Background:} Stein variational gradient descent (SVGD) and its variants have shown promising successes in approximate inference for complex distributions.  In practice, we notice that the kernel used in SVGD-based methods has a decisive effect on the empirical performance. Radial basis function (RBF) kernel with median heuristics is a common choice in previous approaches, but unfortunately this has proven to be sub-optimal. 

\textbf{Method:} Inspired by the paradigm of Multiple Kernel Learning (MKL), our solution to this flaw is using a combination of multiple kernels to approximate the optimal kernel, rather than a single one which may limit the performance and flexibility. Specifically, we first extend Kernelized Stein Discrepancy (KSD) to its multiple kernels view called Multiple Kernelized Stein Discrepancy (MKSD) and then leverage MKSD to construct a general algorithm Multiple Kernel SVGD (MK-SVGD). Further, MK-SVGD can automatically assign a weight to each kernel without any other parameters, which means that our method not only gets rid of optimal kernel dependence but also maintains computational efficiency.

\textbf{Experimental Results:} Experiments on various tasks and models demonstrate that our proposed method consistently matches or outperforms the competing methods.
\end{abstract}

\section{Introduction}
\label{intro}
Bayesian inference is a powerful mechanism aiming at systematically reasoning under uncertainty in machine learning, with broad applications to various areas that involve probabilistic modeling. The core concern of Bayesian inference is the posterior distribution of model parameters, which is yet typically computationally intractable for complex models. Hence, enormous approaches have emerged for approximating posterior distribution, roughly including two branches, sampling-based approaches (e.g., Markov Chain Monte Carlo (MCMC)~\cite{neal2011mcmc,hoffman2014no,zhang2019cyclical}), and optimization-based approaches (e.g.,Variational Inference (VI)~\cite{kingma2013auto,blei2017variational}). With the asymptotic convergence guaranteed, MCMC directly approximates the posterior distribution with the drawn samples but converges slowly in practice due to undesirable auto-correlation between samples. By contrast, VI uses a simpler distribution for approximation by minimizing their Kullback-Leibler (KL) divergence, which is efficient and suitable for large dataset scenarios. Under this circumstance, the pre-defined form of the simple distribution greatly affects the accuracy and computational cost of the approximation. 

Recently, one combines the advantages of sampling-based and optimization-based approaches, yielding a novel approximation method called Stein variational gradient descent (SVGD)~\cite{liu2016stein}. It can be either considered as a particle-based sampling approach~\cite{chen2018unified,liu2018accelerated,zhang2020stochastic,zhang2021dpvi} or as a non-parametric variational inference algorithm~\cite{liu2019understanding}. Specifically, SVGD applies an iteratively transformed set of particles to fit the posterior distribution by minimizing the KL divergence, which is performed with deterministic functional gradient descent. The finally obtained particles can be viewed as samples drawn from the posterior distribution. This method does not require explicit selection of a specific parameter family, thus circumventing the problem of balancing accuracy and computational cost. Meanwhile, since SVGD conducts with deterministic gradient updates, it converges fast and has a series of variants~\cite{han2018stein,detommaso2018stein,wang2019stein}. 

Despite both theoretical guarantees and practically fast, we notice that the performance of SVGD and its variants crucially depends on the choice of the kernel. First, the kernel used in SVGD guides the transformed direction of the particles, which refers to the direction of the steepest descent in the local average derived by weighting the contribution of all particles~\cite{gorham2017measuring}. Then, it also can capture the underlying geometry of the target distribution, making flow the particles along with the support of the target distribution~\cite{hofmann2008kernel}. Hence, the kernel should be specified cautiously and vary according to tasks. However, we find that SVGD and most of its variants generally choose the Radial Basis Function (RBF) with median heuristics as the kernel~\cite{liu2016stein,han2020stein,wang2019stein}. As a commonly used kernel, RBF is simple but has limited expression ability. The current heuristic median method is not good enough at choosing the proper bandwidth for a kernel. And we also show the performance of SVGD with RBF is unstable as shown in Figure~\ref{Fig2.main} of the experimental section. Therefore, to better play the role of the kernel function in SVGD and its variants, some more complex kernels need to be considered.

To this end, we propose a new general algorithm called Multiple Kernel SVGD (MK-SVGD), which is inspired by the paradigm of multiple kernel learning. We first introduce a new discrepancy definition named Multiple Kernelized Stein Discrepancy (MKSD), extending the Kernelized Stein Discrepancy (KSD) to multi-kernel form. Then, based on the MKSD, we present the flexible algorithm MK-SVGD, where the optimal kernel is approximated by weighted multiple kernels instead of a single one. Importantly, the optimal weight of each kernel is learned automatically without additional parameters. It means that MK-SVGD not only improves the ability of expression but also holds computational efficiency.

In summary, our main contributions are as follows:
\begin{itemize}
    \item We introduce a new discrepancy definition named MKSD, which can be regarded as a multi-kernel version of KSD. When using the optimal weight vector, we obtain its salable variant called maxMKSD.

    \item We propose a general algorithm MK-SVGD based on the defined MKSD. MK-SVGD uses a combined kernel for approximation, where each kernel is assigned a weight to measure its significance. This can better capture the underlying geometry structure of the target distribution. In addition, the optimal weight of each kernel is learned automatically in MK-SVGD.
    
    \item We evaluate the MK-SVGD on both synthetic and real-world data of several different tasks, including Multivariate Gaussian, Bayesian logistic regression and Bayesian neural networks. Experimental results have demonstrated the effectiveness of our proposed method.

\end{itemize}

The rest of the paper is organized as follows. We first briefly review the KSD and SVGD in Section~\ref{background}. Then, we present the proposed method MKSD and MK-SVGD in Section~\ref{MK_SVGD}. After that, we introduce some related research in Section~\ref{realted_works}, and the experiments are performed in Section~\ref{experiments}. Finally, we summarize our work and point out the direction of future work in Section~\ref{conclusion}.

\section{Backgrounds}
\label{background}

For a better statement and explanation, we briefly introduce some concepts and algorithms, namely Kernelized Stein Discrepancy (KSD) and Stein variational gradient descent (SVGD), which are the basis of our proposed method. Note that, vectors are denoted by bold lowercase letters; matrices are denoted by upper-case letters; sets are denoted by calligraphic letters throughout this paper. 

\textbf{Preliminary} For any $\alpha_i \in {\mathbb{R}}$ , $\bm{x}_i \in {\mathbb{R}^d}$ and a symmetric function $k:\mathbb{R}^d\times \mathbb{R}^d \rightarrow \mathbb{R}$, if we have $\sum_{ij} \alpha_i k(\bm{x}_i, \bm{x}_j)\alpha_j \ge 0$, then $k$ is a positive definite kernel.  A \textit{Reproducing Kernel Hilbert Space} (RKHS) $\mathcal{H}_k$ is associated with the specific positive definite kernel $k(\bm{x},\bm{x'})$, and consists the closure of functions of form
\begin{align}
    f(\bm{x})=\sum_{i} \alpha_{i} k\left(\bm{x}, \bm{x}_{i}\right), \quad \forall \alpha_{i} \in \mathbb{R}, \quad \bm{x}_{i} \in \mathbb{R}^{d}.
    \label{b_eq1}
\end{align}
Let $g = \sum_j\beta_j k(\bm{x},\bm{x}_j)$, the equipped inner product of the RKHS $\mathcal{H}_k$ is $\langle f, g \rangle_{\mathcal{H}}=\sum_{i j} \alpha_{i} \beta_{j} k\left(\bm{x}_i,\bm{x}_j\right)$ and norm is $\| f \|_{\mathcal{H}}^2 = \sum_{ij} \alpha_i \alpha_j k(\bm{x}_i,\bm{x}_j)$. See \cite{liu2016kernelized} or \cite{berlinet2011reproducing} for more details.

\textbf{Kernelized Stein Discrepancy} All following advanced approaches start from Stein Methods \cite{stein1972bound,barbour2014steins} and RHKS theory. Stein Methods provides a series of tools to bound the distance between two probability distributions with respect to a specific probability metric. Targeting KSD, we would like to review the whole inference process beginning from \textit{Stein Identity}. Let $p(\bm{x})$ be a probability density function, which is positive and continuously differentiable supported on $\mathbb{R}^d$. \textit{Stein Identity} says that for any smooth vector function $\bm{\phi}(\bm{x}) = \left[\phi_{1}(\bm{x}) ,..., \phi_{d}(\bm{x})\right]^{\top}$ we have
\begin{align}
    \mathbb{E}_p \left[\mathcal{A}_p \bm{\phi}(\bm{x})\right]=\mathbb{E}_p \left[\bm{s}_{p}(\bm{x})\bm{\phi}(x)^{\top} + \nabla_{\bm{x}} \bm{\phi}(\bm{x})\right]=0,
    \label{stein_identity}
\end{align}
where $\bm{s}_{p}(\bm{x})=\nabla_{\bm{x}} \log p(\bm{x})$ called \textit{Score function} and $\mathcal{A}_p$ is a linear functional operator on $\bm{\phi}$ called \textit{Stein operator}.
Assuming $\lim _{\|\bm{x}\| \rightarrow \infty} p(\bm{x}) \bm{\phi}(\bm{x})=0$, we can easily check the correctness using integration by parts. With the holding of \textit{Stein Identity}, we call that $\bm{\phi}$ is in the \textit{Stein class} of function $p$. 

Now we can introduce another probability density function $q(\bm{x})$, which is also positive and continuously differentiable supported on $\mathbb{R}^d$ like $p(\bm{x})$. If we directly replace the first $p(\bm{x})$ in Eq.~{\ref{stein_identity}} with $q(\bm{x})$, say consider $\mathbb{E}_q[\mathcal{A}_p \bm{\phi}(\bm{x})]$ where the expectation of $\mathcal{A}_p \bm{\phi}(\bm{x})$ is now under $q(\bm{x})$, then the \textit{Stein Identity} would no longer hold for most general function $\bm{\phi}$. Formally, this can be easily checked by
\begin{align}
    \mathbb{E}_{q}\left[\mathcal{A}_{p} \bm{\phi}(\bm{x})\right]=\mathbb{E}_{q}\left[\left(\bm{s}_{p}(\bm{x})-\bm{s}_{q}(\bm{x})\right) \bm{\phi}(x)^{\top}\right],
    \label{stein_diff}
\end{align}
where $\bm{s}_{p}$ and $\bm{s}_{q}$ are score functions of $p(\bm{x})$ and $q(\bm{x})$ respectively. We notice that Eq.~\ref{stein_diff} no longer equals $0$ unless $\bm{s}_{p} = \bm{s}_q$, which implies $p=q$. This gives us the possibility to measure the difference between two distributions. Given a proper function set $\mathcal{F}$, \textit{Stein Divergence} (SD) measures the difference between $p(\bm{x})$ and $q(\bm{x})$ as
\begin{align}
        \mathbb{D}_{\mathcal{F}}(q \| p) \coloneqq \max _{\bm{\phi} \in \mathcal{F}}\left\{\mathbb{E}_{\bm{x} \sim q}\left[\mathcal{A}_{p} \bm{\phi}(\bm{x})\right]\right\},
        \label{stein_divergence}
\end{align}
which means that the \textit{Stein Divergence} is the maximum violation of \textit{Stein Identity} in function set $\mathcal{F}$. Therefore how to choose a proper function set $\mathcal{F}$ is vital. Traditional methods are generally difficult to optimize or require special considerations \cite{gorham2017measuring1}. Conversely, Kernelized Stein Discrepancy (KSD) solves this problem by taking function set $\mathcal{F}$ to the unit ball of an RKHS, in which the optimization has a closed-form solution. Specifically, KSD is defined as
\begin{align}
        \mathbb{S}(q \| p) \coloneqq \max _{\bm{\phi} \in \mathcal{H}^d}\left\{\mathbb{E}_{\bm{x} \sim q}\left[\mathcal{A}_{p} \bm{\phi}(\bm{x})\right],  \quad \text{s.t.} \quad ||\bm{\phi}||_{\mathcal{H}^d} \le 1\right\}
        \label{KSD}
\end{align}
where $\mathcal{H}$ is a RKHS associated with a kernel $k(\bm{x},\bm{x}')$, which is in the Stein class of $p(x)$. According to \cite{liu2016stein} and \cite{liu2016kernelized}, the optimal solution of Eq.~\ref{KSD} is 
\begin{align}
    \bm{\phi}(\bm{x})=\bm{\phi}^{*}(\bm{x}) /\left\|\bm{\phi}^{*}\right\|_{\mathcal{H}^{d}},\quad \text{where,} \quad \bm{\phi}^{*}(\cdot)=\mathbb{E}_{\bm{x} \sim q}\left[\mathcal{A}_{p} k(\bm{x}, \cdot)\right].
    \label{best_phi}
\end{align}
Substitute the optimal solution into Eq.~\ref{KSD}, the $\mathbb{S}(q \| p)$ is  $\left\|\phi^{*}\right\|_{\mathcal{H}^{d}}^{2}$.

\textbf{Stein Variational Gradient Descent} Let $p(\bm{x})$ defined in above be our target distribution. \textit{Stein Variational Gradient Descent} (SVGD) tends to approximate $p(\bm{x})$ with a set of particles $\{\bm{x_i}\}_{i=1}^n$ in the sense that
\begin{align}
    \mathbb{E}_{p}[f(\bm{x})] = \lim _{n \rightarrow \infty} \frac{1}{n} \sum_{i=1}^{n} f \left(\bm{x}_{i}\right),
\end{align}
where $f$ can be any test function. To achieve this, SVGD randomly samples a collection of particles from an initial simple distribution (e.g. \textit{Standard normal distribution}), and then iteratively updates them by a deterministic map
\begin{align}
    T(\bm{x}_{i}) = \bm{x}_{i}+\epsilon \bm{\phi}^{*}\left(\bm{x}_{i}\right), \quad \forall i=1, \cdots, n,
    \label{particles_map}
\end{align}
where $\bm{\phi}^*:\mathbb{R}^d\rightarrow \mathbb{R}^d$ is the perturbations driving the particles to cover the target distribution $p(\bm{x})$ and $\epsilon$ is a small step size. Assuming $\bm{z}=T(\bm{x})$ as $\bm{x}\sim q(\bm{x})$, the optimal perturbation $\bm{\phi}^*$ can be obtained by solving the optimization problem as follow,
\begin{align}
        \bm{\phi}^{*}=\mathop{\arg \max }_{\bm{\phi} \in \mathcal{F}}\left\{-\left.\frac{d}{d \epsilon} \mathrm{KL}\left(q_{{[T]}} \| p\right)\right|_{\epsilon=0}\right\},
        \label{Opt_phi}
\end{align}
where $\mathcal{F}$ is candidate perturbation function set. The key observation is that 
\begin{align}
    \begin{aligned}
        -\left.\frac{\mathrm{d}}{\mathrm{d} \epsilon} \mathrm{KL}\left(q_{[T]} \| p\right)\right|_{\epsilon=0} 
        &= \nabla_{\bm{x}} \log p(\bm{x}) \bm{\phi}(\bm{x})^{\top}+\nabla_{\bm{x}} \bm{\phi}(\bm{x}) \\
        &= \mathbb{E}_{\bm{x} \sim q}\left[\mathcal{A}_{p} \bm{\phi}(\bm{x})\right],
        \label{key_obs}
    \end{aligned}
\end{align}
which is exactly the objective function in Eq.~\ref{stein_diff}. Following the inference form Eq.~\ref{stein_diff} to Eq.~\ref{best_phi}, we can obtain the optimal perturbations by restricting the set of $\mathcal{F}$ to the unit ball of an RKHS $\mathcal{H}$ as follow
\begin{align}
    \bm{\phi}^{*}(\cdot) \propto \mathbb{E}_{\bm{x} \sim q}[\mathcal{A}_p k(\bm{x}, \cdot)] = \mathbb{E}_{\bm{x} \sim q}\left[\nabla_{\bm{x}} \log p(\bm{x}) k(\bm{x}, \cdot)+\nabla_{\bm{x}} k(\bm{x}, \cdot)\right],
    \label{svgd_opt_phi}
\end{align}
where $k$ is the positive definite kernel associated with RKHS $\mathcal{H}$. 
In practice, SVGD iteratively updates particles $\{\bm{x}_i\}_{i=1}^n$ by $\bm{x}_i\leftarrow \bm{x}_i + \epsilon \bm{\phi}^*(\bm{x}_i)$, where,
\begin{align}
     \bm{\phi}^{*}\left(\bm{x}_{i}\right)=\frac{1}{n} \sum_{j=1}^{n}\left[\nabla_{\bm{x}_{j}} \log p\left(\bm{x}_{j}\right) k\left(\bm{x}_{j}, \bm{x}_{i}\right)+\nabla_{\bm{x}_{j}} k\left(\bm{x}_{j}, \bm{x}_{i}\right)\right].
     \label{svgd_update}
 \end{align}
The different parts in the Eq.~\ref{svgd_update} play different roles: the first part drives the particles to the high probability of target distribution, while the second part acts as a \textit{repulsive force} to separate different particles to cover the entire density area.

\section{SVGD with Multiple Kernel}
\label{MK_SVGD}
In this section, we introduce our proposed new method step by step in the following order: Firstly, we intuitively analyze the role of the kernel in Kernelized Stein Discrepancy (KSD) and its importance. Then combined with Multiple Kernel Learning (MKL), we propose a new discrepancy definition to address these issues, named Multiple Kernelized Stein Discrepancy (MKSD). After that, we leverage MKSD to construct a general algorithm based on SVGD, which be called Multiple Kernel SVGD (MK-SVGD). Finally, we simply analyze the time consumption of our algorithm. 
\subsection{Intuition}
Before moving on to more details, we first give an intuitive analysis of the kernel selection issue in KSD and its variants. We notice that the role of the kernel is mainly reflected in the following two aspects: guiding the particles to the direction of the steepest descent in the local average by weighting the contribution of each particle~\cite{gorham2017measuring} and flowing the particles along with the support of the target distribution~\cite{hofmann2008kernel}. Therefore, selecting a suitable kernel is crucial and the optimal kernel varies by task. In practice, RBF kernel with median heuristics is a common choice but mostly sub-optimal. Some other complex kernels, such as deep kernels~\cite{wilson2016deep}, might be preferred but not easy to use. To this end, we propose a new optimal kernel approximation method based on multiple kernel learning (MKL), which utilizes a more powerful combined kernel instead of a single one. We also introduce a weight for each kernel to measure its importance. It is worth noting that the weights can be automatically assigned without additional parameters. In short, our method not only gets rid of optimal kernel dependence but also maintains computational effectiveness. 

\subsection{Multiple Kernelized Stein Discrepancy}
For a better elaboration, we first briefly introduce the basic concepts of Multiple Kernel Learning~(MKL). Simply put, MKL~\cite{kang2018self,xu2009extended,xu2010simple,gonen2011multiple,wilson2016deep} provides an automatic way to learn the optimal combination of base kernels. The goal of MKL is to maximize a generalized performance measure by finding an optimal base kernel functions combination. In general, MKL-based methods outperform single-kernel ones due to the better feature representation ability and flexibility~\cite{pan2019self}. The combination way can be roughly divided into three categories according to different tasks and needs: Linear combination, Nonlinear combination, and Data-dependent combination~\cite{gonen2011multiple}. Considering the computational efficiency, we use a linear way that combines the kernels as 
\begin{align}
        k_{\mathbf{w}}(\bm{x},\bm{x}')=\sum_{i}^{m} w_i k_i(\bm{x},\bm{x}'), \quad \text{s.t.} \quad \mathbf{w} \in \mathbb{R}_{+}^m, ||\mathbf{w}||_{2}=1
        \label{eq11}
\end{align}
where $m$ is the number of base kernels and $\mathbf{w}$ is the weight vector. Referring to Section~\ref{background}, we have known that the kernel in KSD or SVGD is required to be in the Stein class for smooth densities in a proper sense. Some commonly used kernels meet the requirements, such as RBF kernel $k(\bm{x},\bm{x}')=\text{exp}(-\frac{1}{h}\|\bm{x} - \bm{x}'\|_2^2)$. Therefore, it is first necessary to check the suitability of the multi-kernel, that is, whether it belongs to the Stein class for smooth densities in $\mathcal{X}=\mathbb{R}^d$. 

\begin{prop}(Multi-kernel in Stein class)
    Assume a sequence of kernels $k_i(\bm{x},\bm{x}')$ are in the Stein class, then we have $k_{\mathbf{w}}(\bm{x},\bm{x}')=\sum_i \bm{w}_{i} k_{i}(\bm{x},\bm{x}')$ is in the Stein class for any $\mathbf{w} \in \mathbb{R}_+^m$. See proof in~\ref{proof_of_mk_stein_class}.
    \label{prop MK_sc}
\end{prop}

According to \textit{Proposition~\ref{prop MK_sc}}, we know that the multi-kernel is eligible as long as its base kernels are in Stein class of smooth densities in $\mathcal{X}=\mathbb{R}^d$ and the weight vector $\mathbf{w}$ is non-negative. This means that the correctness of our algorithm has a theoretical guarantee. Now, it is natural to take advantage of the multi-kernel to extend the generalization of KSD. Specifically, we propose a new discrepancy based on KSD by directly replacing a single kernel with a multi-kernel, called Multiple Kernelized Stein Discrepancy~(MKSD). The formal definition is as follows:
\begin{definition}
    (Multiple Kernelized Stein Discrepancy) Assume $q(\bm{x})$ and $p(\bm{x})$ are continuously differentiable densities supported on $\mathcal{X} \subseteq \mathbb{R}^{D}$ and a sequence of  base kernels $k_i(\bm{x},\bm{x}')$ are in the Stein class of $q(\bm{x})$. By defining
    \begin{align}
        u_{p}^{i}(\bm{x},\bm{x}') = &\bm{s}_p(\bm{x})^\top k_{i}(\bm{x},\bm{x}')\bm{s}_p(\bm{x}') + 
        \bm{s}_p(\bm{x})^\top \nabla_{\bm{x}'}k_{i}(\bm{x},\bm{x}') \notag \\ 
        &+\nabla_{\bm{x}}k_{i}(\bm{x},\bm{x}')^\top   \bm{s}_q(\bm{x}') + \operatorname{Tr}(\nabla_{\bm{x},\bm{x}'}k_{i}(\bm{x},\bm{x}')) ,\notag
    \end{align}
    where $\operatorname{Tr}$ represents the trace of the matrix, $\bm{s}_p$ and $\bm{s}_q$ are score function of $p(\bm{x})$ and $q(\bm{x})$ respectively,  then we have
    \begin{align}
        u_{p}^{\mathbf{w}}(\bm{x},\bm{x}') 
        &= \bm{s}_p(\bm{x})^\top k_{\mathbf{w}}(\bm{x},\bm{x}') \bm{s}_p(\bm{x}') +  \bm{s}_p(\bm{x})^\top \nabla_{\bm{x}'}k_{\mathbf{w}}(\bm{x},\bm{x}') \notag\\
        & \quad~  + \nabla_{\bm{x}}k_{\mathbf{w}}(\bm{x},\bm{x}')^\top   \bm{s}_q(\bm{x}') + \operatorname{Tr}(\nabla_{\bm{x},\bm{x}'}k_{\mathbf{w}}(\bm{x},\bm{x}')) \notag\\
        &=\sum_i^m w_i u_{p}^{i}(\bm{x},\bm{x}'). \notag
    \end{align}
    Finally, we reach the definition of MKSD as follows:
    \begin{align}
        \mathbb{S}_{k_{\mathbf{w}}}(q \| p) &= \mathbb{E}_{\bm{x},\bm{x}'\sim q} [u_{p}^{\mathbf{w}}(\bm{x},\bm{x}')] \notag
        =\sum_i^m w_i \mathbb{E}_{\bm{x},\bm{x}'\sim q} [u_{p}^{i}(\bm{x},\bm{x}')] \notag\\
        &= \sum_i^m w_i \mathbb{S}_{k_i}(q \| p).
        \label{Eq_MKSD}
    \end{align}
    where $\mathbb{S}_{k_i}(q \| p)$ is the KSD corresponding to the base kernel $k_i$.
    \label{definition: MKSD}
\end{definition}
From Eq.~\ref{Eq_MKSD} in \textit{Definition~\ref{definition: MKSD}}, we note that MKSD is defined as the linear combination of KSDs corresponding to the base kernel. In addition, the weight of each KSD is exactly the weight of the corresponding base kernel in the multi-kernel. The advantage of this definition is straightforward to extend while maintaining efficiency. To consider MKSD as a discrepancy metric, however, we need some more qualitative verification, which is done in \textit{Proposition~\ref{prop: MKSD as a discrepancy}}. Moreover, a corollary about the maximum MKSD is given in Corollary~\ref{corollary: maxMKSD}.
\begin{prop}
    (MKSD as Discrepancy) Assume a sequence $k_i(\bm{x},\bm{x}')$ are integrally strictly positive definite, and $q$, $p$ are continuous densities with $\mathbb{E}_q(\bm{x})[u_{p}^{\mathbf{w}}(\bm{x},\bm{x}')] < \infty$ for all $\mathbf{w} \in \mathbb{R}_{+}^m$, 
    we have $\mathbb{S}_{k_{\mathbf{w}}}(q \| p) \geq 0 $ and $\mathbb{S}_{k_{\mathbf{w}}}(q \| p) = 0$ if and only if $q =p$. See proof in~\ref{proof_MKSD_as_discrepancy}.
\label{prop: MKSD as a discrepancy}
\end{prop}

\begin{corollary}
    (maxMKSD) Assume the conditions in Definition~\ref{definition: MKSD} are satisfied. Then
    \begin{align}
        \text{MKSD}_{max}(q,p) = \max_{\mathbf{w} \in \mathbb{R}_{+}^m, ||\mathbf{w}||_{2}=1} \mathbb{S}_{k_{\mathbf{w}}}(q \| p)
    \end{align}
    is equal to 0 if and only if $p$ = $q$ a.e.
    \label{corollary: maxMKSD}
\end{corollary}

\subsection{SVGD with Multiple Kernels}
As mentioned in Eq.~\ref{svgd_update}, the role of the kernel in guiding particle update in SVGD is mainly reflected in the following two aspects: Pushing each particle towards the high probability areas of target distribution by weighting the gradient direction of each particle in the first term; Acting as a \textit{repulsive force} that spreads the particle along with target distribution and doesn't collapse into the local modes in the second terms. Take RBF kernel $k(\bm{x},\bm{x'})=\text{exp}(-\frac{1}{h}\|\bm{x}-\bm{x'}\|_2^2)$ as an example. The corresponding \textit{repulsion force} term is $\mathbb{E}_{\bm{x'}\sim q} [\nabla_{\bm{x'}} k(\bm{x},\bm{x'})] = \mathbb{E}_{\bm{x'}\sim q} [ \frac{2}{h}(\bm{x}-\bm{x}')k(\bm{x},\bm{x}')]$, driving point $\bm{x}$ away from its neighboring point $\bm{x'}$ who has a large $k(\bm{x}, \bm{x'})$. Besides, the bandwidth $h$ plays its role in turning the smoothness of repulsive force term, e.g. the repulsive term vanishes when $h \rightarrow 0$. Therefore, if bandwidth $h$ can be adjusted according to the current surrounding of particles, the kernel could better capture the underlying geometry structure of the target distribution $p(\bm{x})$, then each particle can efficiently spread the target distribution. 

To this end, we propose a more general algorithm, Multiple Kernel SVGD (MK-SVGD). The key idea is to combine the different kernels with self-adjusted weights to approximate the optimal kernel, and thus get rid of the dependence on the single kernel which may limit the performance and flexibility. As a result, our algorithm can dynamically perform kernel selection during SVGD updates. Next, we will introduce MK-SVGD in detail.

Let $\bm{\phi}_{\mathbf{w}}:\mathbb{R}^d\rightarrow \mathbb{R}^d$ be a perturbation function in the unit ball of a general RKHS $\mathcal{H}_{k_{\mathbf{w}}}$ equipped with the multiple kernels $k_{\mathbf{w}}(\bm{x}, \bm{x}')=\sum_i^m w_i k_i(\bm{x}, \bm{x}')$. Then moving the particles through the transition function $T_{\mathbf{w}}(\bm{x})=\bm{x}+\epsilon \bm{\phi}_{\mathbf{w}}(\bm{x})$. Assuming $q_{[T_{\mathbf{w}}]}(\bm{z})$ as the density of $\bm{z} = T_{\mathbf{w}}(\bm{x})$ when $\bm{x} \sim q(\bm{x})$, the steepest descent direction $\bm{\phi}_{\mathbf{w}}^{*}$ can be obtained by solving the optimization problem as follow,
\begin{align}
        \bm{\phi}_{\mathbf{w}}^{*}=\mathop{\arg \max }_{\bm{\phi}_{\mathbf{w}} \in \mathcal{H}_{k_{\mathbf{w}}}}\left\{-\left.\frac{\mathrm{d}}{\mathrm{d} \epsilon} \mathrm{KL}\left(q_{[T_{\mathbf{w}}]} \| p\right)\right|_{\epsilon=0}\right\}.
        \label{Opt_phi_w}
\end{align}
Similar to Eq.~\ref{key_obs}, we have
\begin{align}
    \begin{aligned}
        -\left.\frac{\mathrm{d}}{\mathrm{d} \epsilon} \mathrm{KL}\left(q_{[T_{\mathbf{w}}]} \| p\right)\right|_{\epsilon=0} 
        &= \nabla_{\bm{x}} \log p(\bm{x}) \bm{\phi}_{\mathbf{w}}(\bm{x})^{\top}+\nabla_{\bm{x}} \bm{\phi}_{\mathbf{w}}(\bm{x}) \\
        &= \mathbb{E}_{\bm{x} \sim q}\left[\mathcal{A}_{p} \bm{\phi}_{\mathbf{w}}(\bm{x})\right] \\
        &= -\mathbb{S}_{k_{\mathbf{w}}}(q \| p).
        \label{key_obs_w}
    \end{aligned}
\end{align}
From Eq.~\ref{key_obs_w}, we know that the gradient of KL divergence respect to $\epsilon$ in Eq.~\ref{Opt_phi_w} is exactly the negative MKSD. Therefore, with weight vector $\mathbf{w}$, the steepest descent direction is
\begin{align}
    \bm{\phi}^{\star}_{\mathbf{w}}(\cdot) = \mathbb{E}_{\bm{x}\sim q} [\mathcal{A}_p k_{\mathbf{w}}(\bm{x},\cdot)]=\sum_i w_i \mathbb{E}_{\bm{x}\sim q} [\mathcal{A}_p k_{i}(\bm{x},\cdot)]=\sum_i w_i\bm{\phi}^{\star}_i(\cdot),
    \label{Opt_phi_w_star}
\end{align}
where $\bm{\phi}^{\star}_i(\cdot)$ denotes the optimal direction related to $i$-th base kernel. In other words, given a weight vector w, the optimal update direction is jointly determined by the  steepest descent direction of each base kernel. In practice, MK-SVGD iteratively updates particles by $\bm{x}\leftarrow \bm{x} + \epsilon \bm{\phi}_{\mathbf{w}}^*(\bm{x})$, where,
\begin{align}
     \bm{\phi}_{\mathbf{w}}^{*}\left(\bm{x}\right)=\sum_{i=1}^{m}\frac{w_i}{n} \sum_{j=1}^{n}\left[\nabla_{\bm{x}_{j}} \log p\left(\bm{x}_{j}\right) k_{i}\left(\bm{x}_{j}, \bm{x}\right)+\nabla_{\bm{x}_{j}} k_{i}\left(\bm{x}_{j}, \bm{x}\right)\right].
     \label{mk-svgd_update}
 \end{align}
 
 Next, we focus on weight vector $\mathbf{w}$. As mentioned before, the weight vector is vital to choosing the final direction for each particle by linearly combining the base kernels. In the different particles surrounding, the optimal combination of base kernels is also different, so we need to update the weight vector as the particles move. Formally, we need to find a set of weights that maximize MKSD, which is exactly the problem explained in Corollary \ref{corollary: maxMKSD}. Therefore, the problem can be turned into an optimization problem about $\mathbf{w}$, as 
\begin{align}
    \mathbf{w}^{\star}=\underset{\mathbf{w}}{\arg \max } ~ \mathbb{S}_{k_{\mathbf{w}}}\left(q \| p\right) = \underset{ \mathbf{w}}{\arg \max } \big\langle w_i, \mathbb{S}_{k_i}(q \| p) \big\rangle.
    \label{opt_w}
\end{align}
This is a straightforward optimization problem and the maximum is achieved when
\begin{align}
    w_i = \sqrt{\mathbb{S}_{k_i}(q \| p)/\sum_i {\mathbb{S}_{k_i}(q \| p)}} = ||\bm{\phi}_{i}||_{\mathcal{H}_d} / \sqrt{\sum_i ||\bm{\phi}_{i}||^2_{\mathcal{H}_d}}.
    \label{opt_w_i}
\end{align}

Combining all the above steps results in a more general and powerful algorithm, Multiple Kernel Stein Variational Gradient Descent (MK-SVGD), which is clearly summarised in \textit{Algorithm~\ref{alg: MK_SVGD}}. 
\begin{algorithm}[htbp]
    \SetKwInOut{Input}{Input}
    \SetKwInOut{Output}{Output}
    \SetKwInOut{Output1}{Output1}
    
    \SetAlgoLined
    \Input {Particles $\{\bm{x}_i\}_{i=1}^n$ initial from simple prior distribution;\\  
            A sequence kernel function $\{k_i\}_{i=1}^m$; \\
            Weight vector $\{w_i\}_{i=1}^{m}$ initialized with $\frac{1}{m}$;\\
            Step size $\epsilon$; Iteration number $T$.
            }
    \Output{Final particles $\{\bm{x}_i\}_{i=1}^n$ that represents the target distribution $p(\bm{x})$}
    \For{$t \leq T$}{
    
        \textbf{Update particles} \\
        \quad Update all particles $\{\bm{x}_i\}_{i=1}^n$ by $\bm{x}_i\leftarrow \bm{x}_i + \epsilon \bm{\phi}_{\mathbf{w}}^*(\bm{x}_i)$ and calculate $\bm{\phi}_{\mathbf{w}}^*(\cdot)$ using Eq.~\ref{mk-svgd_update}
            
        \textbf{Update weight vector $\mathbf{w}$}\\
        \quad Update weight vector $\{w_i\}_{i=1}^{m}$ using Eq.~\ref{opt_w_i}
    }
    \caption{Multiple Kernel Stein Variational Gradient Descent}
    \label{alg: MK_SVGD}
\end{algorithm}

Finally, we simply analyze the time complexity of our algorithm. From \cite{liu2016stein}, we know that the major computation bottleneck in iterative procedure of SVGD lies on calculating the gradient $\nabla_{\bm{x}}\text{log}\  p(\bm{x})$ for all the points $\{ \bm{x}_i \}_{i=1}^n $. Therefore, the introduction of multi-kernel in our algorithm does not significantly increase the time consumption. For example, if we use multiple RBF kernels $k(\bm{x},\bm{x'})=\text{exp}(-\frac{1}{h}\|\bm{x}-\bm{x'}\|_2^2)$ with different bandwidth $h$, the term $\| \bm{x} - \bm{x'} \|_2^2$ is shared by each kernel. In other words, we only need to compute the rest part for the kernels with different $h$, and the time consumption is linearly increased compared to the SVGD, which is $\mathcal{O}(mn^2)$. Additional speedup method, such as mini-batch, also can be obtained similar to SVGD.

\section{Related Works}
\label{realted_works}
As a pioneering work, SVGD is a very attractive algorithm. Therefore, many exciting and effective works have been proposed based on SVGD in the machine learning community. Generally speaking, the works related to SVGD are mainly concentrated in two research lines. One is to combine SVGD with other models or embed SVGD into other models, such as~\cite{feng2017learning,pu2017vae,li2017gradient,korba2020non,liu2021profiling}. Another major concern is the improvement or generalization of the vanilla SVGD algorithm, such as~\cite{han2018stein,detommaso2018stein,wang2019stein,chen2020projected,jaini2021learning,ba2021understanding}. Next, we will briefly introduce several representative works in these two directions.

In the first case, SVGD is often used as a part of the model. For example, in work \cite{feng2017learning}, SVGD was used to train stochastic neural networks with the purpose of drawing samples from the target distribution for probability inference. Based on SVGD, a new learning method for variational auto-encoders~(VAEs) is proposed in \cite{pu2017vae}.  Another work~\cite{li2017gradient} proposes an algorithm to directly estimate the score function of the implicitly deﬁned distribution, which is similar to the main framework of SVGD.

Our work is an extension of the vanilla SVGD, which belongs to the second case. There also exist many works in the second case. Using a surrogate gradient to replace true gradient in SVGD, \cite{han2018stein} proposes a gradient-free SVGD that focuses on the target distributions with intractable gradient. Besides, in work~\cite{han2020stein}, the authors extend the idea of gradient-free SVGD to discrete distribution and solved the limitation that original SVGD can only work under continuous distribution. Another work~\cite{detommaso2018stein} accelerates and generalized the SVGD algorithm to approximate a Newton-like iteration in function space by including second-order information. The Scaled Hessian kernel is proposed using second-order information. Furthermore, the work \cite{wang2019stein} leverages a more general matrix-valued kernel to generate SVGD for ﬂexibly incorporating preconditioning information. Note that, Our method is intrinsically different from~\cite{wang2019stein}. The work~\cite{wang2019stein} introduces the optimization problem to vector-valued RKHS with matrix-valued kernels, while our method focuses on leveraging the effectiveness of  Multiple Kernel Learning, which optimizes the objective in real-valued RKHS.

\section{Experiments}
\label{experiments} 
To evaluate the performance of MK-SVGD, we conduct sufficient experiments on both synthetic and real-world data of several different tasks. Firstly, we test our algorithm on a multivariate gaussian distribution as a toy example and then move on to a series of more challenging tasks concluding Bayesian logistic regression and Bayesian neural networks. For a fair comparison, we adopt basic experimental settings similar to~\cite{liu2016stein}. We select RBF kernels with different bandwidths as the basic components of the multi-kernel. In specific, to provide a wider range of scales, we use 10 RBFs with a ratio of 0.1 between each adjacent bandwidth $h$. Besides, the bandwidth range varies for different experiments, and the specific settings are described in each experiment section. All particles are randomly sampled from the prior unless otherwise specified. The step size updating is controlled by AdaGrad~\cite{duchi2011adaptive}.
The comparison methods involved in the experiment are summarized as follows,
\begin{itemize}
    \item \textit{Probabilistic back-propagation~(PBP)}~\cite{hernandez2015probabilistic}, a classic scalable method for learning Bayesian neural networks. It is also the only non-SVGD-based method in the comparison methods, using the implementation by~\cite{liu2016stein};
    \item \textit{Vanilla SVGD}, the baseline of SVGD-based methods and code is also available on~\cite{liu2016stein};
    \item \textit{Matrix-SVGD}, a recent advanced SVGD-method use the constant preconditioning matrix kernel, using the code in~\cite{wang2019stein};
    \item \textit{MK-SVGD}, Multiple Kernel SVGD, our new propose method.
\end{itemize}

\subsection{Multivariate Gaussian}
\label{multi-gaussian}
We start our experiments with a toy example in simulating a multivariate Gaussian distribution. Given a target distribution
\begin{align}
        p(x) = \mathcal{N}(x;\mu,\Sigma),\quad \text{where}\quad \mu = [-0.6871, 0.8010],\ \Sigma= \begin{bmatrix} 0.2260 & 0.1652 \\ 0.1652 & 0.6779 \end{bmatrix},
\end{align}
we use MK-SVGD to get a set of particles that cover the target distribution nicely. Note that the parameters of target distribution settings are randomly chosen. In detail, we firstly initialize 500 particles randomly sampling from a standard normal distribution $\mathcal{N}(x;0, I)$ which is far away from the target distribution. Then the particles are updated through MK-SVGD by 200 epochs. In addition, the bandwidths of 10 RBF kernels are in the range of: $[2^{-4}, 2^{-3},\cdots,2^5]$. 

\begin{figure}[H]
    \centering  
    \subfigure[2D example]{
    \label{Fig1.sub.1}
    \includegraphics[width=0.47\textwidth]{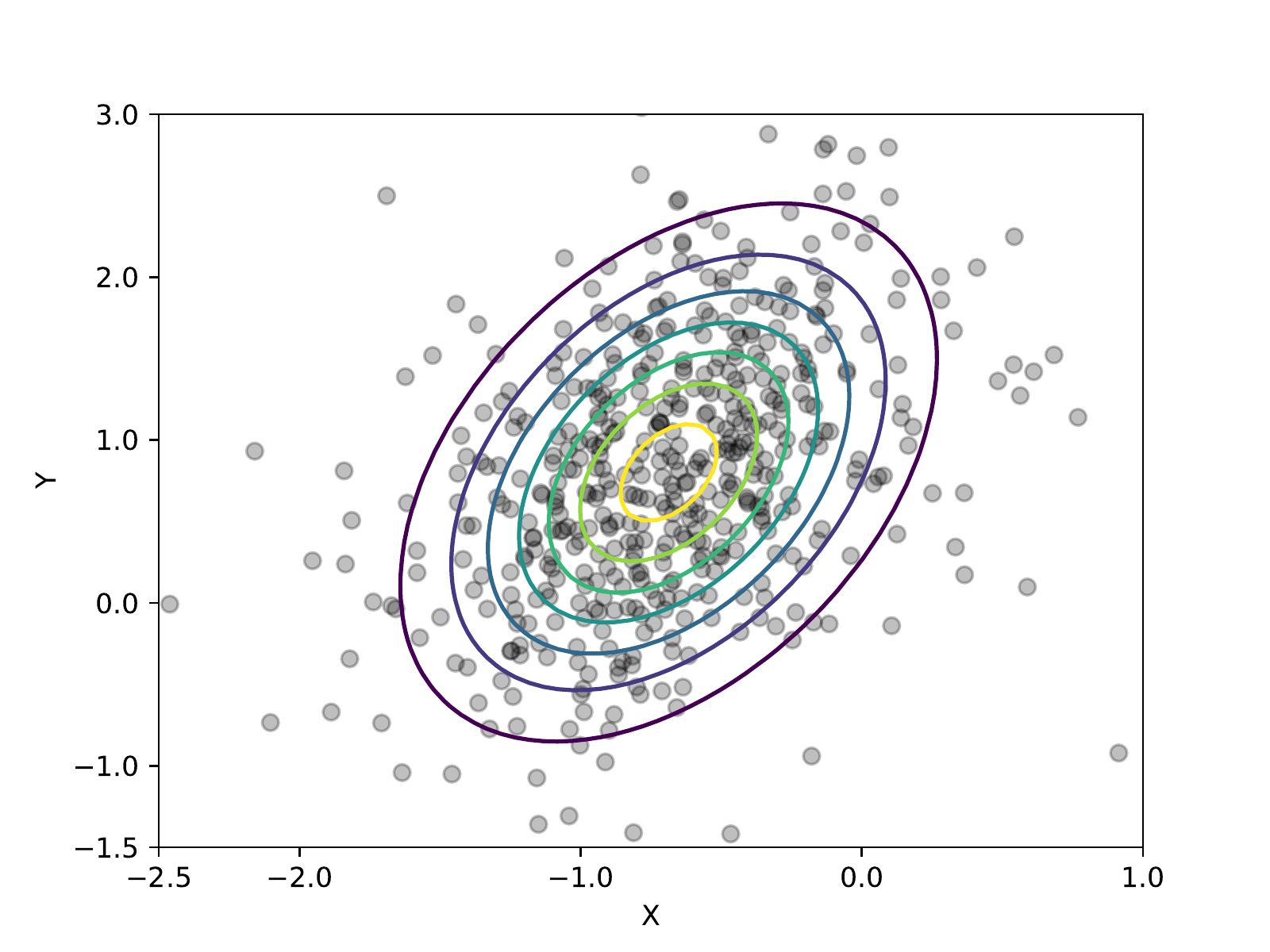}}
    \subfigure[Weight of each kernel]{
    \label{Fig1.sub.2}
    \includegraphics[width=0.47\textwidth]{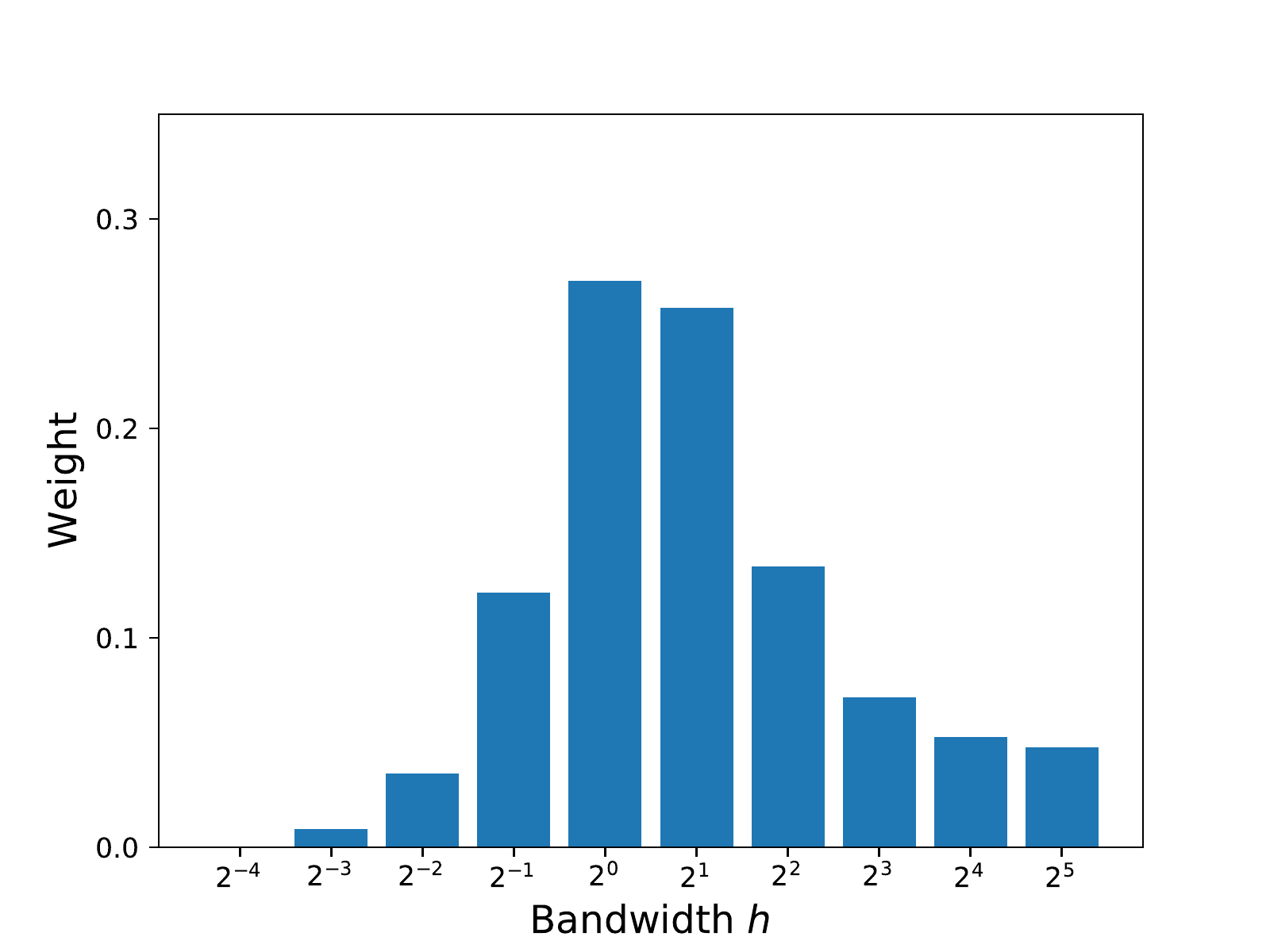}}
    \caption{Results of multivariate Gaussian. (a)~The final updated particles, which fit target distribution well. (b)~Weight distribution corresponding to each base kernel.}
    \label{Fig1.main}
\end{figure}

\paragraph{Results} After particles updating, we get the final set of particles, which can be viewed as sampling samples from the target distribution. Figure~\ref{Fig1.sub.1} shows the final results, from which we can see that the particle group fits the target distribution well. According to the weight distribution of each kernel shown in Figure~\ref{Fig1.sub.2}, we find that the kernels with bandwidth $h=1$ and $h=2$ play a major role. Further, we calculate the mean of each variable of the final particle group over ten random runs, and the value is $[-0.68792629, \ 0.80107447]$, which is pretty close to the groundtruth.

\subsection{Bayesian Logistic Regression}
We then evaluate MK-SVGD on more complex models and real data. As for the model, we firstly consider a binary classification problem based on the Bayesian logistic regression. Denote the dataset as $\bm{D} = \{ (\bm{x}_i, y_i) \}_{i=1}^N$, where $\bm{x}_i$ is the feature vector, $y_i\in \{0,1\}$ is the corresponding binary label and $N$ is the number of data points. Then a Bayesian logistic regression model for binary classiﬁcation defines as follow
\begin{align}
    p(\bm{\theta} \mid \bm{D}) &\propto p(\bm{D} \mid \bm{\theta}) p(\bm{\theta})
\end{align}
with
\begin{align}
    p(\bm{D} \mid \bm{\theta}) &= \prod_{j=1}^{N}\left[y_{j} \sigma\left(\bm{\theta}^{\top} \bm{x}_{j}\right)+\left(1-y_{j}\right) \sigma\left(-\bm{\theta}^{\top} \bm{x}_{j}\right)\right],
\end{align}
where $\sigma(*)$ is a standard logistic function, defined as $\sigma(*)\coloneqq 1/(1+\text{exp}(-*))$, and the prior of $\theta$ is a Gaussian distribution $p_0(\bm{\theta}|\alpha)=\mathcal{N}(\bm{\theta};0,\alpha^{-1})$ with $p(\alpha)=Gamma(\alpha;a,b)$.
The value of the hyperparameters $a$ and $b$ in Gamma distribution are set to 0 and 0.01 respectively. Our interest is using MK-SVGD to generate a set of samples $\{ \bm{\theta} \}_{i=1}^n$ to approximate the posterior distribution $p(\bm{\theta} \mid \bm{D})$. After that, we can use the particles group to predict the class labels of test data points. 

As for the dataset, we use the \textit{Covtype}\footnote{https://www.csie.ntu.edu.tw/~cjlin/libsvmtools/datasets/binary.html}, which is a binary dataset with 581,012 data points, and each data point has 54 features. We select 20\% of the dataset for testing and the rest for training. For all training process, we use AdaGrad optimizer with a learning rate of 0.05 for both $\bm{\theta}$ and $\mathbf{w}$. The mini-batch size is 100. We use $n=100$ particles for all methods. Unlike toy example in section~\ref{multi-gaussian}, we select bandwidth of 10 RBF kernels in the range of: $[2^{2}, 2^{3},\cdots,2^{11}]$. All results are averaged on 10 random trials.
\begin{figure}[H]
    \centering  
    \subfigure[Testing Accuracy]{
    \label{Fig2.sub.1}
    \includegraphics[width=0.49\textwidth]{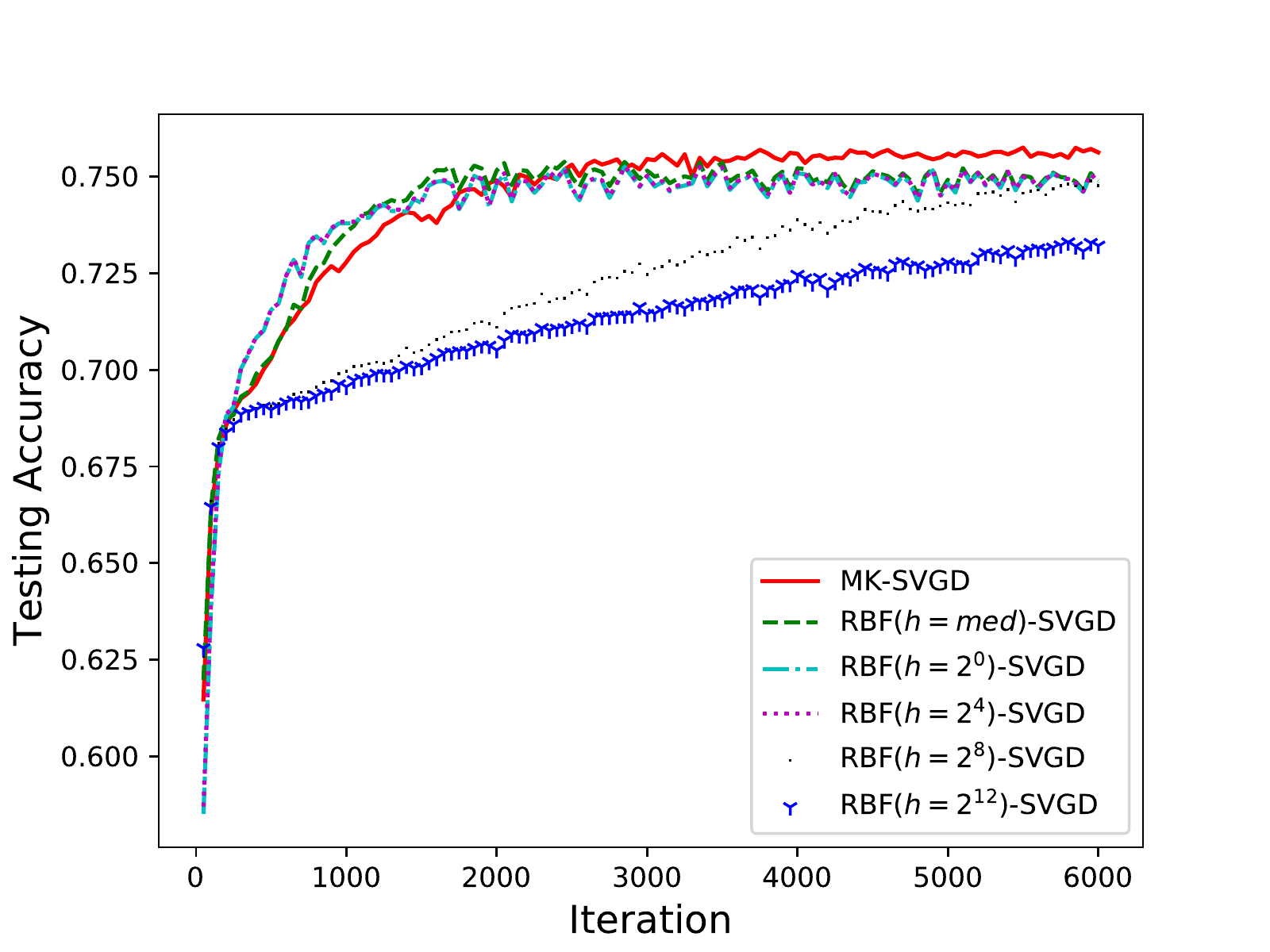}}
    \subfigure[Testing Log-likelihood]{
    \label{Fig2.sub.2}
    \includegraphics[width=0.49\textwidth]{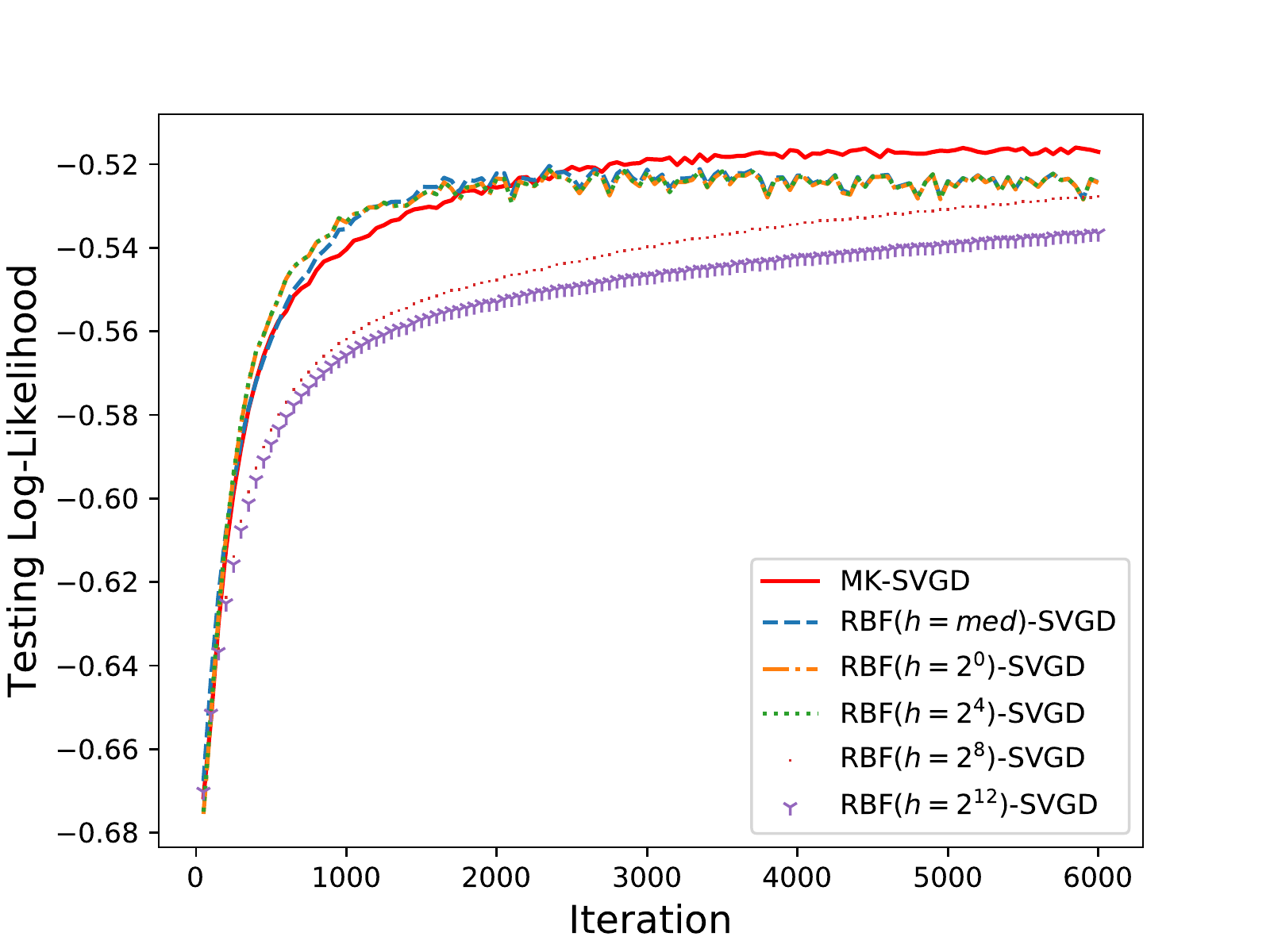}}
    \caption{Bayesian Logistic regression results of SVGD with different kernel on the Covtype dataset.}
    \label{Fig2.main}
\end{figure}
\paragraph{Results} Figure~\ref{Fig2.main} show the results of MK-SVGD and vanilla SVGD with different kernels in two aspects: Test accuracy and Test log-likelihood. Specifically, we compare MK-SVGD with five vanilla SVGDs with different bandwidths, which are $h=2^0$, $h=2^4$, $h=2^8$, $h=2^{12}$ and median heuristics way. We can see that the final results of our algorithm are better than vanilla SVGD on both test accuracy and test log-likelihood. Moreover, it can be clearly seen that our algorithm results are more stable. While the convergence speed is slightly slower due to the mixture process of multi-kernel.

\subsection{Bayesian Neural Networks}

Finally, we test our algorithm on Bayesian neural network regression on several UCI datasets\footnote{https://archive.ics.uci.edu/ml/datasets.php}. All datasets information used in our experiments are summarized in Table~\ref{tab:dataset_intro}. For all datasets, we use neural network with one hidden layer with 50 units and ReLU activation functions. We initialize the network weights using isotropic Gaussian prior and do not scale the input of the output layer. We perform 10 random trials for all results and then take the average. As the experimental setup remains the same, the results for PBP are obtained directly from~\cite{liu2016stein}. All methods use $n=20$ particles. All datasets are randomly selected 90\% for training and the rest are used for testing. For the weight of kernel, we use gradient descent or AgaGrad for different datasets. The mini-batch size is 100. In addition, we select bandwidth of 10 RBF kernels in the range of: $[2^{-4}, 2^{-3},\cdots,2^5]$.
\begin{table}[!htbp]
        \centering
        \resizebox{\textwidth}{!}{
        \begin{tabular}{l|cccc}
            \toprule
             \textbf{Dataset} & \textbf{PBP} & \textbf{Vanilla SVGD} & \textbf{Matrix-SVGD} & \textbf{MK-SVGD}\\
             \midrule
             Boston & $2.977 \pm 0.093$ & $2.957 \pm 0.099$ & $2.898 \pm 0.184$ & $\bm{2.750 \pm 0.316}$ \\
             Combined & $4.052 \pm 0.031$ & $4.033 \pm 0.033$ & $4.056 \pm 0.033$ & $\bm{4.029 \pm 0.110}$\\
             Concrete & $5.506 \pm 0.103$ & $5.324 \pm 0.104$ & $\bm{4.869 \pm 0.124}$ & $5.162 \pm 0.219$\\
             Wine & $0.614 \pm 0.008$ & $0.609\pm0.010$ & $0.637 \pm 0.008$ &$\bm{0.601\pm0.050}$ \\
             Yacht & $0.778 \pm 0.042$ & $0.864 \pm 0.052$ & $2.750\pm 0.125$ & $\bm{0.688 \pm 0.100}$\\
             \bottomrule
        \end{tabular}}
        \caption{Average Test RMSE.}
        \label{tab:test_rmse}
    \end{table}
    
    \begin{table}[!htbp]
        \centering
        \resizebox{\textwidth}{!}{
        \begin{tabular}{l|cccc}
            \toprule
             \textbf{Dataset} & \textbf{PBP} & \textbf{Vanilla SVGD} & \textbf{Matrix-SVGD} & \textbf{MK-SVGD}\\
             \midrule
             Boston & $-2.579 \pm 0.052$ & $-2.504 \pm 0.029$ & $-2.669 \pm 0.141$ & $\bm{-2.474 \pm 0.070}$\\
             Combined & $-2.819 \pm 0.008$ & $\bm{-2.815 \pm 0.008}$ & $ -2.824 \pm 0.009 $ & $-2.835 \pm 0.035$\\
             Concrete & $-3.137 \pm 0.021$ & $-3.082 \pm 0.018$ & $-3.150 \pm 0.054$ & $\bm{-3.080 \pm 0.031}$\\
             Wine & $-0.931 \pm 0.014$ & $-0.925\pm0.014$ & $-0.980\pm0.016$ & $\bm{-0.902\pm0.040}$ \\
             Yacht & $-1.221\pm0.044$ &$-1.225 \pm 0.042$ & $ -2.390\pm 0.452$ & $\bm{-1.199 \pm 0.053}$\\
             \bottomrule
        \end{tabular}}
        \caption{Average Log-Likelihood.}
        \label{tab:test_ll}
    \end{table}
    
\paragraph{Results} We compare MK-SVGD with vanilla SVGD, probabilistic back-propagation (PBP) and Matrix-SVGD in terms of the Root Mean Square Error(RMSE) and the log-likelihood on test data. All results are shown in Table~\ref{tab:test_rmse} and Table~\ref{tab:test_ll} respectively. We can see our algorithm yields better performance than other methods in most cases.

\section{Conclusion}
\label{conclusion}
In this work, we define a new discrepancy metric MKSD, and then propose a more general and efficient algorithm MK-SVGD based on MKSD. Taking advantage of MKL, MK-SVGD gets rid of the dependence on the optimal kernel choice in SVGD-based methods with single kernel. Moreover, MK-SVGD can automatically learn a weight vector to adjust the importance of each base kernel without any other parameters, which guarantees computational efficiency. Experiments on different data and models show the effectiveness of our algorithm.

\begin{acknowledgements}
This paper was supported by the National Key Research and Development Program of China (No.~2018AAA0100204), and a key program of fundamental research from Shenzhen Science and Technology Innovation Commission (No.~JCYJ20200109113403826).
\end{acknowledgements}

\bibliographystyle{spmpsci}      
\bibliography{ref}   

\newpage
\appendix

\section{Definitions}
\label{appendix_a}

\begin{definition}(Strictly Positive Kernel) For any function $f$ that satisfies $0 \le \|f\|_2^2 \le \infty$, a kernel $k(x,x')$ is said to be integrally strictly positive definition if
    \begin{align}
        \int_{\mathcal{X}} f(x)k(x,x')f(x') dx dx' > 0.
        \notag
    \end{align}
    
    \label{definition: positive kernel}
\end{definition}

\begin{definition} (Stein Class)
    \label{stein_class}
    For a smooth function $f: \mathcal{X} \rightarrow \mathbb{R}$, then, we can say that $f$ is in the Stein class of $q$ if satisfies
    $$
        \int_{x \in \mathcal{X}} \nabla_{x}(f(x) q(x)) d x=0
    $$
\end{definition}

\begin{definition}(Kernel in Stein Class)
    If kernel $k\left(x, x^{\prime}\right)$ has continuous second order partial derivatives, and for any fixed $x$, both $k(x, \cdot)$ and $k(\cdot, x)$ are in the Stein class of $p$, then the kernel $k\left(x, x^{\prime}\right)$ said to be in the Stein class of $p$.
    \label{define_s}
\end{definition}

\section{Proof}
\label{appendix_b}

\paragraph{\textbf{Proof of Proposition~\ref{prop MK_sc}}}
    \begin{proof}
        For the kernel $k_{\mathbf{w}}(x,x')=\sum_{i=1}^m w_i k_i(x,x')$, where $\mathbf{w} \in \mathbb{R}_{+}^m, ||\mathbf{w}||_{2}=1$, and $k_i(x,x')$ is in the Stein class. According to \ref{define_s}, we know that $k_i(x,x')$ has continuous second order partial derivatives.\\
        Setting $g_i$ be the continuous second order partial, we know that the continuous second order partial $g$ of multiple kernel $k_{\mathbf{w}}$ is
        \begin{align}
            g = \sum_{i=1}^{m} w_i g_i
        \end{align}
        \notag
        So, the kernel $k_{\mathbf{w}}(x,x')=\sum_{i=1}^m w_i k_i(x,x')$, where $\mathbf{w} \in \mathbb{R}_{+}^m, ||\mathbf{w}||_{2}=1$, is in the Stein class.
        \label{proof_of_mk_stein_class}
    \end{proof}

\paragraph{\textbf{Proof of Proposition \ref{prop: MKSD as a discrepancy}}}
    \label{proof_MKSD_as_discrepancy}
    \begin{proof}
        We know $\mathbb{S}_{k_i}(q\|p) \geq 0$ and $\mathbb{S}_{k_i}(q\|p) = 0$ if and only if $q=p$ a.e. This means $\mathbb{S}_{k_{\mathbf{w}}} \geq 0$ and $\mathbb{S}_{k_{\mathbf{w}}}(q,p) = 0$ when $q = p$ a.e. for any $\mathbf{w} \in \mathbb{R}_{+}^{m}$.
        Then, we know $\mathbb{S}_{k_{\mathbf{w}}}(q,p) = 0$ if and only if $\mathbb{S}_{k_i}(q,p) = 0$. Thus $q = p$ a.e.
    \end{proof}

\section{Datasets}
\begin{table}[!htbp]
    \centering
    \begin{tabular}{c|c|c}
        \toprule
         \textbf{Dataset} & \textbf{Number of Attributes} & \textbf{Number of Instances} \\
         \midrule
         Covtype & 54 & 581012 \\
         \midrule
         Boston & 14 & 506 \\
         Combined & 5 & 9568 \\
         Concrete & 10 & 1030 \\
         Wine & 12 & 4898 \\
         Yacht & 7 & 308 \\
         \bottomrule
    \end{tabular}
    \caption{Datasets used in our experiments. Covtype, used in Bayesian Logistic Regression; Other datasets, used in Bayesian Neural Networks.}
    \label{tab:dataset_intro}
    
\end{table}
\end{document}